\documentclass[10pt,twocolumn,letterpaper]{article}

\usepackage{cvpr}
\usepackage{times}
\usepackage{epsfig}
\usepackage{graphicx}
\usepackage{amsmath}
\usepackage{amssymb}

\usepackage{algorithmicx,algorithm}

\usepackage{multirow}
\usepackage{makecell}
\usepackage{microtype}
\usepackage{subfigure}
\usepackage{booktabs}

\usepackage[breaklinks=true,bookmarks=false]{hyperref}

\cvprfinalcopy % *** Uncomment this line for the final submission

\def\cvprPaperID{****} % *** Enter the CVPR Paper ID here

\begin{document}

%%%%%%%%% TITLE
\title{Ensemble Models with Batch Spectral Regularization and Data Blending for Cross-Domain Few-Shot Learning with Unlabeled Data}

\author{Zhen Zhao, Bingyu Liu, Yuhong Guo, Jieping Ye\\
AI Tech, DiDi ChuXing
}

\maketitle
\thispagestyle{empty}

%%%%%%%%% ABSTRACT
\begin{abstract}

In this paper, we present our proposed ensemble model with batch spectral regularization and data blending mechanisms for the Track 2 problem of the 
	cross-domain few-shot learning (CD-FSL) challenge. 
We build a multi-branch ensemble framework by using diverse feature transformation matrices,
while deploying batch spectral feature regularization on each branch to improve the model's transferability. 
Moreover, we propose a data blending method to exploit the unlabeled data 
	and augment the sparse support set in the target domain.
Our proposed model demonstrates effective performance 
	on the CD-FSL benchmark tasks. 
\end{abstract}

%%%%%%%%% BODY TEXT
\section{Introduction}

Despite the success of deep neural networks on large-scale labelled data, 
their performance degrades severely 
on datasets with only a few labeled instances. 
In order to advance the research progress in this area, the Cross-Domain Few-Shot Learning (CD-FSL) challenge ~\cite{guo2019new} has been initiated, where miniImageNet is used as the source domain, and four other datasets 
including the plant disease images (CropDiseases~\cite{mohanty2016using}), satellite images (EuroSAT~\cite{helber2019eurosat}), dermoscopic images of skin lesions (ISIC2018~\cite{tschandl2018ham10000,codella2019skin}), and X-ray images (ChestX~\cite{wang2017chestx}) are used as the target domains. In this challenge, we tackle the task of track2, cross-domain few-shot learning with unlabeled data, 
where a separate unlabeled subset in each target domain can be used during training.

One key challenge of the cross-domain few-shot learning task lies in the large cross-domain gaps, 
which limits the effective generalization ability of classic few-shot learning methods. 
In this paper, we propose to use
a batch spectral feature regularization (BSR) mechanism 
to prevent over-fitting of the prediction model to the source domain
and increases the generalization capacity of the model across large domain gaps. 
Moreover, we deploy a feature transformation based ensemble model 
that constructs and learns multiple prediction networks in multiple diverse feature spaces to improve the model's robustness. 
To exploit the unlabeled data in the target domain,
we propose a data blending strategy 
that combines the unlabeled data and the support set to augment the sparse labeled support instances during a fine-tuning stage in the target domain. 
We also apply the idea of label propagation~\cite{ye2017labelless} to refine the classification results on the target query instances.
Our overall model demonstrates effective performance on the track 2 CD-FSL tasks. 

%%%%%%%%%%%%%%%%%%%%%%%%%%
\section{Proposed Method}

\begin{figure*}[htbp]
\begin{center}
   \includegraphics[width=0.9\linewidth]{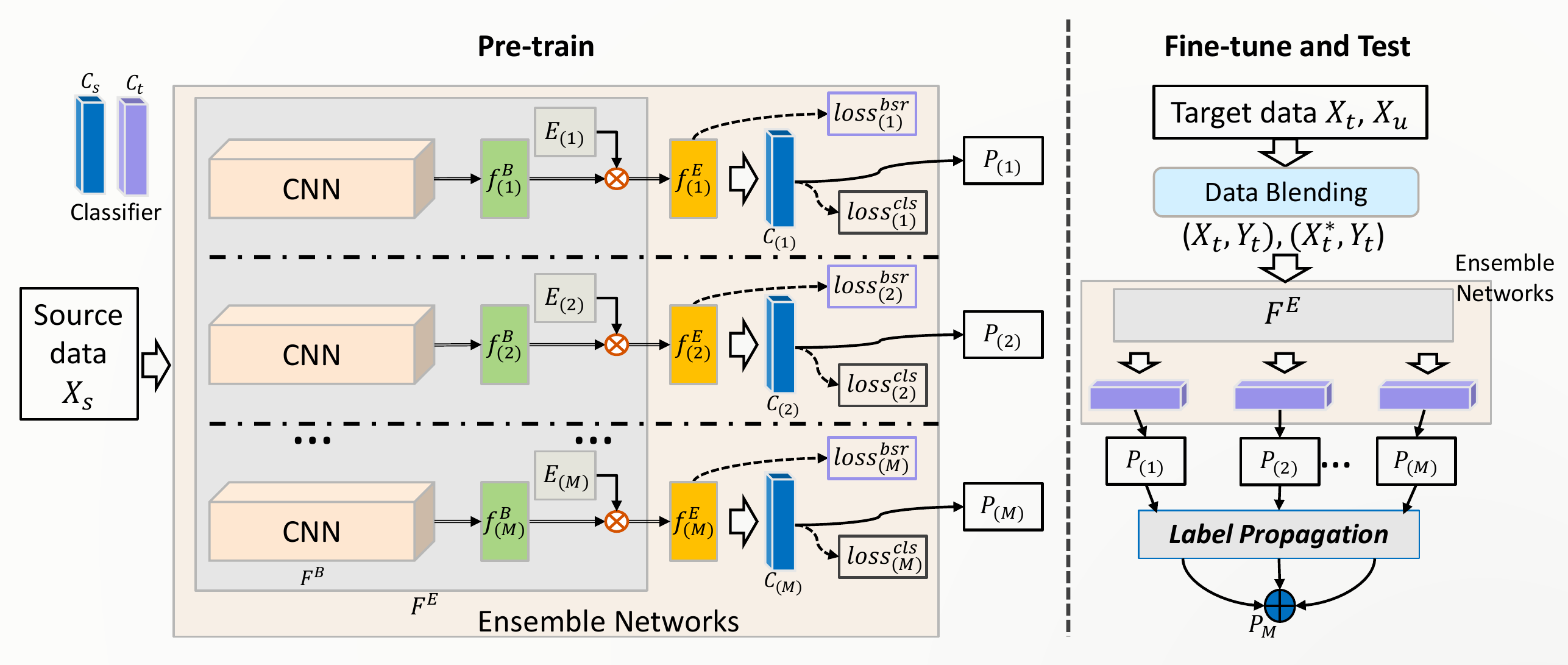}
\end{center}
   \caption{An overview of the proposed approach.}
\label{fig:overview}
\end{figure*}

We consider the following problem setting.
We have a set of labeled images $(X_s,Y_s)$ in the source domain. 
In the target domain, we have a set of few-shot learning tasks.
In each task, we have a labeled support set,
$(X_t,Y_t)$, which contains a few (N) labeled images from each of the K classes (K-way N-shot),
and a set of query images, $X_q$, that are used to evaluate a CD-FSL method's performance.
In addition, we also have a set of unlabeled data, $X_u$, in the target domain.
Specifically, we follow the learning setting and evaluation strategy in~\cite{guo2019new}. 
The overall architecture of our proposed model is illustrated in Fig.~\ref{fig:overview}.
We present each component of the model below.

\subsection{Ensemble Model}

As shown in Fig.~\ref{fig:overview}, we build an ensemble model with multiple prediction branch networks
in diverse feature spaces. 
We increase the diversity of the feature spaces of the model by 
applying a different feature projection matrix $E$ on each branch network. 
With $M$ branches (e.g., $M=10$), we use $M$ randomly generated symmetric matrices $\{Z_{(1)},Z_{(2)},...,Z_{(M)}\}$ 
to produce $M$ orthogonal matrices $\{E_{(1)},E_{(2)},...,E_{(M)}\}$, where each $E_{(i)}$ is the orthogonal matrix 
composed of the eigenvectors of $Z_{(i)}$. The feature vector $f^B_{(i)}$ extracted by the convolutional neural network $F_B$ can be transformed to a new feature representation vector via $f^E_{(i)}=E_{(i)}f^B_{(i)}$, and then sent to the classifier $C_{(i)}$ for classification. 
The $M$ diverse models can be trained in the source domain using the $M$ generated orthogonal matrices. 
The training of the network is conducted by minimizing a standard cross-entropy loss. The batch-wise loss function for a single network can be written as:
\begin{equation}
L_{cls} = \frac{1}{b}\sum\limits_{j = 1}^b {{L_{ce}}\left( {C_i(f_i^E(x_j)),{y_j}} \right)}
\end{equation}
where $L_{ce}$ is the cross entropy loss function, and $b$ is the current batch size.

\subsection{Batch Spectral Regularization}

Inspired by ~\cite{chen2019catastrophic}, we introduce a batch spectral regularization (BSR) mechanism to suppress all singular values of the batch feature matrix during pre-training, which can avoid the problem of overfitting to the source domain. In a single model, given a batch $(X_B, Y_B)$ with size $b$, and its feature matrix $A=[f^E_1,...,f^E_b]$ can be obtained. The BSR can then be written as:
\begin{equation}
{L_{bsr}(A)} = \sum\limits_{i = 1}^b{{\sigma}^2_i}
\end{equation}
where ${\sigma}_1,{\sigma}_2,...,{\sigma}_b$ are the singular values of the batch feature matrix A. The spectral regularized training loss for each batch will be:

\begin{equation}
L = L_{cls}+{\lambda}{L_{bsr}}
\end{equation}

\subsection{Data Blending}

Inspired by the mixup methodology~\cite{zhang2017mixup}, 
we introduce a data blending strategy 
to exploit the information of the additional unlabeled data in the target domain 
and improve the model performance. 
Specifically, we propose to generate a pseudo support set using the mixup method.
For a pair of random instances $[(X^t_i,Y^t_i),X^u_j]$ in a given mini-batch, 
where $X^t_i$ and $X^u_j$ are the instances from the support set and the unlabeled set respectively, we can create a new data instance
\begin{equation}
X^{t*}_i =(1-w)X^t_i+wX^u_j 
\end{equation}
where $w\in[0,1]$ is parameter controls the degree of blending, and obtain a trainable ``pseudo-labeled'' instance $(X^{t*}_i,Y^t_i)$, 
which maintains the information from both the support set and the unlabeled set. 
By fine-tuning the pre-trained prediction model on the support set and the new generated data together, 
we expect to enhance the robustness and capacity of the prediction network. The fine-tuning loss function with data blending can be written as :
\begin{equation}
L_{ft} = L_t(X^t,Y^t)+{\mu}L^*_t(X^{t*},Y^t)
\end{equation}
where $L_t$ and $L^*_t$ are the cross-entropy losses on the support set and the generated pseudo support set respectively, and ${\mu}$ is a trade-off parameter.

In addition to the components above, we also apply a label propagation (LP) procedure to refine the prediction results on the query set, following 
the LP procedure in~\cite{ye2017labelless}.

%%%%%%%%%%%%%%%%%%%%%%%%%%%%%%%%%%%%%%%%%%%%%%%%%%

\begin{table*}[htbp]
\begin{center}
\caption{Results on the CD-FSL benchmark tasks with unlabeled data.}
\resizebox{\textwidth}{!}{
\begin{tabular}{cccc|ccc}
\toprule
\multirowcell{2}{\textbf{Methods}} & \multicolumn{3}{c|}{\textbf{ChestX}} & \multicolumn{3}{c}{\textbf{ISIC}} \\ 
\cline{2-7} 
& 5-way 5-shot & 5-way 20-shot & 5-way 50-shot & 5-way 5-shot & 5-way 20-shot & 5-way 50-shot \\ 
\cline{1-7}
Fine-tuning~\cite{guo2019new} & 25.97\%\(\pm\)0.41\% & 31.32\%\(\pm\)0.45\% & 35.49\%\(\pm\)0.45\% & 48.11\%\(\pm\)0.64\% & 59.31\%\(\pm\)0.48\% & 66.48\%\(\pm\)0.56\%\\
BSDB & 27.50\%\(\pm\)0.45\% & 34.62\%\(\pm\)0.50\% & 37.80\%\(\pm\)0.53\% & 53.52\%\(\pm\)0.63\% & 65.12\%\(\pm\)0.62\% & 70.76\%\(\pm\)0.64\%\\
BSDB+LP & 27.47\%\(\pm\)0.46\% & 34.93\%\(\pm\)0.49\% & 38.46\%\(\pm\)0.55\% & 54.95\%\(\pm\)0.68\% & 66.55\%\(\pm\)0.61\% & 71.87\%\(\pm\)0.62\%\\

\cline{1-7}
BSDB (Ensemble) & 28.38\%\(\pm\)0.47\% & 37.75\%\(\pm\)0.51\% & 42.10\%\(\pm\)0.54\% & 54.54\%\(\pm\)0.65\% & 67.66\%\(\pm\)0.62\% & 74.27\%\(\pm\)0.56\%\\
BSDB+LP (Ensemble) & 28.40\%\(\pm\)0.46\% & 38.17\%\(\pm\)0.53\% & 42.73\%\(\pm\)0.53\% & 56.17\%\(\pm\)0.66\% & 68.95\%\(\pm\)0.60\% & 75.08\%\(\pm\)0.54\%\\

\bottomrule
\end{tabular}}
\resizebox{\textwidth}{!}{
\begin{tabular}{cccc|ccc}
\toprule
\multirowcell{2}{\textbf{Methods}} & \multicolumn{3}{c|}{\textbf{EuroSAT}} & \multicolumn{3}{c}{\textbf{CropDiseases}} \\ 
\cline{2-7} 
& 5-way 5-shot & 5-way 20-shot & 5-way 50-shot & 5-way 5-shot & 5-way 20-shot & 5-way 50-shot \\ 
\cline{1-7}
Fine-tuning~\cite{guo2019new} & 79.08\%\(\pm\)0.61\% & 87.64\%\(\pm\)0.47\% & 90.89\%\(\pm\)0.36\% & 89.25\%\(\pm\)0.51\% & 95.51\%\(\pm\)0.31\% & 97.68\%\(\pm\)0.21\%\\
BSDB & 83.14\%\(\pm\)0.61\% & 90.63\%\(\pm\)0.38\% & 94.03\%\(\pm\)0.28\% & 93.48\%\(\pm\)0.42\% & 98.19\%\(\pm\)0.18\% & 99.20\%\(\pm\)0.11\%\\
BSDB+LP & 85.43\%\(\pm\)0.58\% & 92.30\%\(\pm\)0.34\% & 95.06\%\(\pm\)0.27\% & 95.31\%\(\pm\)0.37\% & 98.90\%\(\pm\)0.16\% & 99.54\%\(\pm\)0.09\%\\

\cline{1-7}
BSDB (Ensemble) & 84.50\%\(\pm\)0.55\% & 92.20\%\(\pm\)0.33\% & 95.17\%\(\pm\)0.25\% & 94.05\%\(\pm\)0.41\% & 98.47\%\(\pm\)0.18\% & 99.30\%\(\pm\)0.10\%\\
BSDB+LP (Ensemble) & 86.66\%\(\pm\)0.54\% & 93.57\%\(\pm\)0.31\% & 96.07\%\(\pm\)0.24\% & 95.93\%\(\pm\)0.37\% & 99.16\%\(\pm\)0.13\% & 99.62\%\(\pm\)0.08\%\\

\bottomrule
\end{tabular}}
\end{center}
\label{table:result}
\end{table*}

\begin{table}[htbp]
\begin{center}
\caption{Average results across all datasets and shot levels.}
\resizebox{0.7\columnwidth}{!}{
\begin{tabular}{c|c}
\toprule
\textbf{Methods} & \textbf{Average}\\ 
\hline
	Fine-tuning~\cite{guo2019new} & 67.23\%\ (0.46\%)\\
	BSDB & 70.67\%\ (0.46\%)\\
	BSDB+LP & 71.73\%\ (0.44\%)\\

\hline
	BSDB (Ensemble) & 72.36\%\ (0.43\%)\\
	BSDB+LP (Ensemble) & 73.38\%\ (0.42\%)\\

\bottomrule
\end{tabular}}
\end{center}
\label{table:avg}
\end{table}
\section{Experiments}
%%%%%%%%%%%%%%%%%%%%%%%%%%%%%%%%%%%%%%%%%%%%%%%%%%

\subsection{Experiment Details}

In the experiments, we follow the protocol of~\cite{guo2019new}, using 15 images of each category as the query set, 
and using 600 randomly sampled few-shot learning tasks in each target domain. The average accuracy and 95$\%$ confidence interval are reported. We use Resnet-10 as the backbone $F_B$ and the fully connected layer as the classifier $C$. In the per-training process, models are trained
for 400 epoch, and the hyperparameter is set as ${\lambda} = 0.001$. The networks are trained by SGD with an initial learning rate of 0.001, a momentum of 0.9, and a weight decay of 0.0005. 
In the fine-tuning process, we set $\mu=0.1$ and $w=0.5$, set the learning rate to 0.01 and conduct fine-tuning for 100 epochs.
In the label propagation step, we use $K=10$ for the K-NN graph construction, set its hyperparameter $\alpha=0.5$. 

\subsection{Experiment Results}
We compare the proposed model with the strong fine-tuning baseline method reported in ~\cite{guo2019new}. 
We report the results of several variants of the proposed model,
including a single prediction network with BSR and Data Blending without ensemble, denoted as {\bf BSDB}, and its extension with the label propagation (LP). 
Then, we extend these variants to the ensemble framework with 10 branches. The CD-FSL results in four different target domains are reported in Table~\ref{table:result}, and the average accuracy across all datasets and shot levels are shown in Table~\ref{table:avg}.

We can see that the single model BSDB has already outperformed the fine-tuning baseline (70.67$\%$ vs 67.23$\%$). 
With the ensemble model with 10 branches, the model performance can be further improved (72.36$\%$). By further adding the LP refinement, we obtain the best performance with BSDB+LP (ensemble) (73.38$\%$). 

\section{Conclusion}
In this paper, we presented an ensemble model with batch spectral regularization and data blending for Cross-Domain Few-Shot Learning With Unlabeled Data. 
In the pre-training process, batch spectral regularization is deployed, and in the fine-tuning process, the information of the unlabeled data 
is exploited through data blending. We also further refined the prediction results with label propagation. 
The overall method exhibited excellent CD-FSL performance.

{\small
\bibliographystyle{ieee_fullname}
\bibliography{egbib}

\begin{thebibliography}{1}\itemsep=-1pt

\bibitem{chen2019catastrophic}
Xinyang Chen, Sinan Wang, Bo Fu, Mingsheng Long, and Jianmin Wang.
\newblock Catastrophic forgetting meets negative transfer: Batch spectral
  shrinkage for safe transfer learning.
\newblock In {\em NeurIPS}, 2019.

\bibitem{codella2019skin}
Noel Codella, Veronica Rotemberg, Philipp Tschandl, M~Emre Celebi, Stephen
  Dusza, David Gutman, Brian Helba, Aadi Kalloo, Konstantinos Liopyris, Michael
  Marchetti, et~al.
\newblock Skin lesion analysis toward melanoma detection 2018: A challenge
  hosted by the international skin imaging collaboration (isic).
\newblock {\em arXiv preprint arXiv:1902.03368}, 2019.

\bibitem{guo2019new}
Yunhui Guo, Noel~CF Codella, Leonid Karlinsky, John~R Smith, Tajana Rosing, and
  Rogerio Feris.
\newblock A new benchmark for evaluation of cross-domain few-shot learning.
\newblock {\em arXiv preprint arXiv:1912.07200}, 2019.

\bibitem{helber2019eurosat}
Patrick Helber, Benjamin Bischke, Andreas Dengel, and Damian Borth.
\newblock Eurosat: A novel dataset and deep learning benchmark for land use and
  land cover classification.
\newblock {\em IEEE Journal of Selected Topics in Applied Earth Observations
  and Remote Sensing}, 12(7):2217--2226, 2019.

\bibitem{mohanty2016using}
Sharada~P Mohanty, David~P Hughes, and Marcel Salath{\'e}.
\newblock Using deep learning for image-based plant disease detection.
\newblock {\em Frontiers in plant science}, 7:1419, 2016.

\bibitem{tschandl2018ham10000}
Philipp Tschandl, Cliff Rosendahl, and Harald Kittler.
\newblock The ham10000 dataset, a large collection of multi-source
  dermatoscopic images of common pigmented skin lesions.
\newblock {\em Scientific data}, 5:180161, 2018.

\bibitem{wang2017chestx}
Xiaosong Wang, Yifan Peng, Le Lu, Zhiyong Lu, Mohammadhadi Bagheri, and
  Ronald~M Summers.
\newblock Chestx-ray8: Hospital-scale chest x-ray database and benchmarks on
  weakly-supervised classification and localization of common thorax diseases.
\newblock In {\em CVPR}, 2017.

\bibitem{ye2017labelless}
Meng Ye and Yuhong Guo.
\newblock Labelless scene classification with semantic matching.
\newblock In {\em BMVC}, 2017.

\bibitem{zhang2017mixup}
Hongyi Zhang, Moustapha Cisse, Yann~N Dauphin, and David Lopez-Paz.
\newblock mixup: Beyond empirical risk minimization.
\newblock {\em arXiv preprint arXiv:1710.09412}, 2017.

\end{thebibliography}
}

\end{document}